\documentclass[conference]{IEEEtran}
\usepackage{times}
\usepackage[numbers]{natbib}
\usepackage{multicol}
\usepackage[bookmarks=true]{hyperref}
\usepackage[nolist]{acronym}

\usepackage{graphicx}
\usepackage{comment}
\usepackage{xcolor}
\usepackage{float}
\usepackage{amsmath}
\usepackage{amssymb}
\usepackage{caption}
\usepackage{subcaption}
\usepackage{todonotes}

\let\vec\boldvec 
\newcommand{\robotloc}{\vec{x}}

\DeclareMathOperator{\rmap}{q}

\begin{document}

\title{Heterogeneous Robot Teams for \\Informative Sampling}

\author{Travis Manderson, Sandeep Manjanna and Gregory Dudek\\ Mobile Robotics Lab, McGill University, Montreal, Canada\\{\tt\small (travism, msandeep, dudek)@cim.mcgill.ca}}



%

\maketitle

\begin{abstract}
In this paper we present a cooperative multi-robot strategy to adaptively explore and sample environments that are unfavorable for humans. We propose a methodology for a team of heterogeneous robots to collaborate on information based planning for applications like sampling thermal imagery in a wildfire affected site to assist with detecting spot fires and areas of residual fires, fire mapping and monitoring fire progression or applications in marine domain for coral reef monitoring and survey. We use Gabor filter based texture classifier on aerial images from an \ac{UAV} to segment the region of interest into classes. A policy gradient based path planner is used on the texture classified aerial image to plan a path for the \ac{UGV}. The \ac{UGV} then uses a local planner to reach the goals set by the global planner by avoiding obstacles. The \ac{UGV} also learns the labels for the segmented classes as drivable and non-drivable using the feedback from the performance while reaching the planned waypoints. We evaluated the building blocks of our approach and present the results with application of these strategies to different domains.
\end{abstract}

\IEEEpeerreviewmaketitle

\section{Introduction}

\begin{figure}[h]
    \begin{center}
    \begin{subfigure}[]{0.25\textwidth}
    	\captionsetup{justification=centering}
        \includegraphics[width=\textwidth]{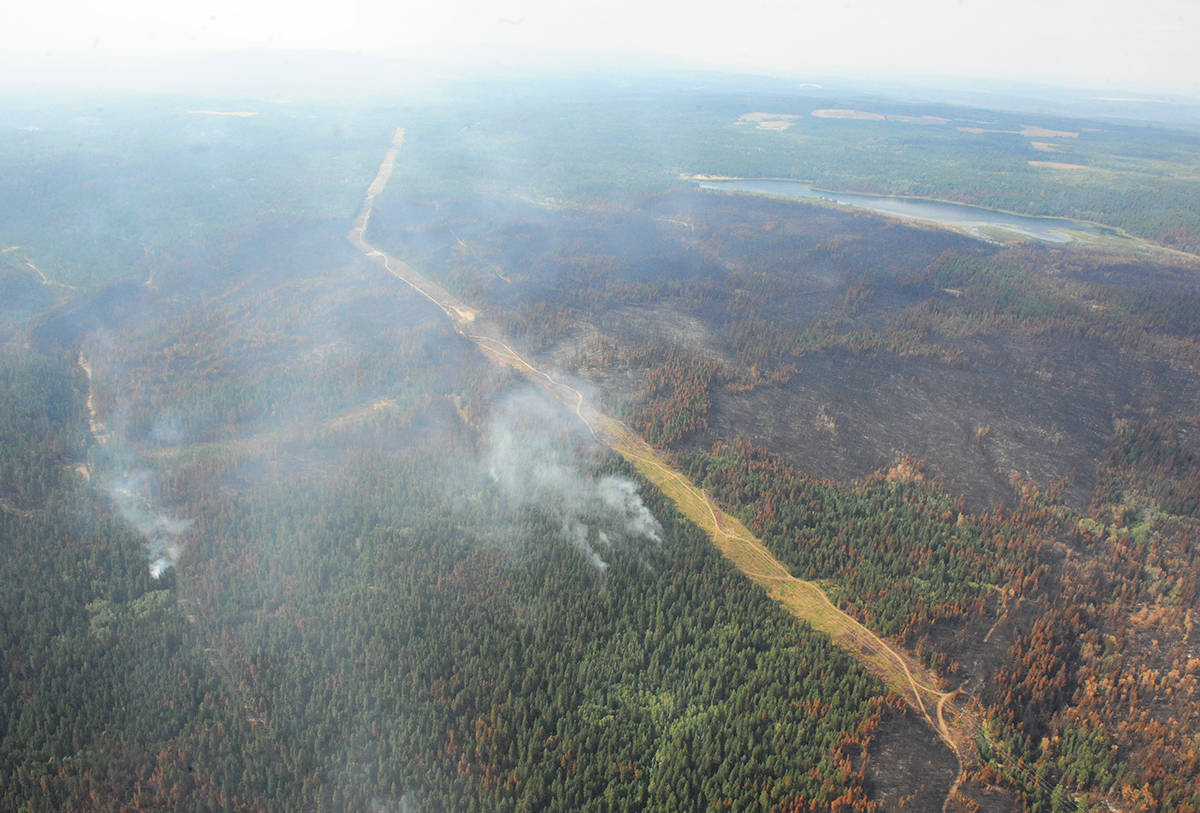}
        \caption{}
        \label{fig:hotspot}
    \end{subfigure}
    \begin{subfigure}[]{0.23\textwidth}
    	\captionsetup{justification=centering}
        \includegraphics[width=\textwidth]{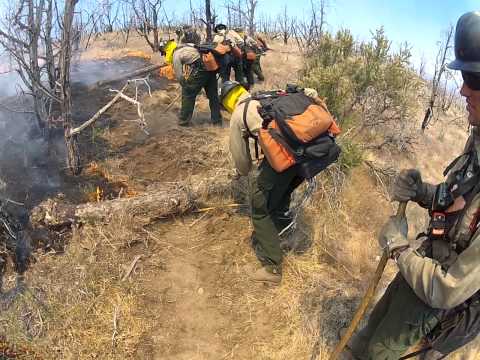}
        \caption{}
        \label{fig:initial_attack}
    \end{subfigure}
    \begin{subfigure}[]{0.25\textwidth}
    	\captionsetup{justification=centering}
        \includegraphics[width=\textwidth]{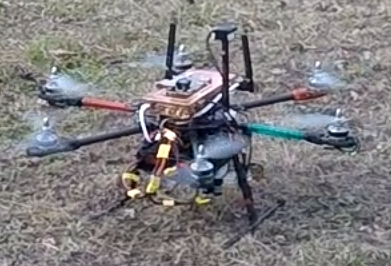}
        \caption{}
        \label{fig:uav}
    \end{subfigure}
    \begin{subfigure}[]{0.225\textwidth}
    	\captionsetup{justification=centering}
        \includegraphics[width=\textwidth]{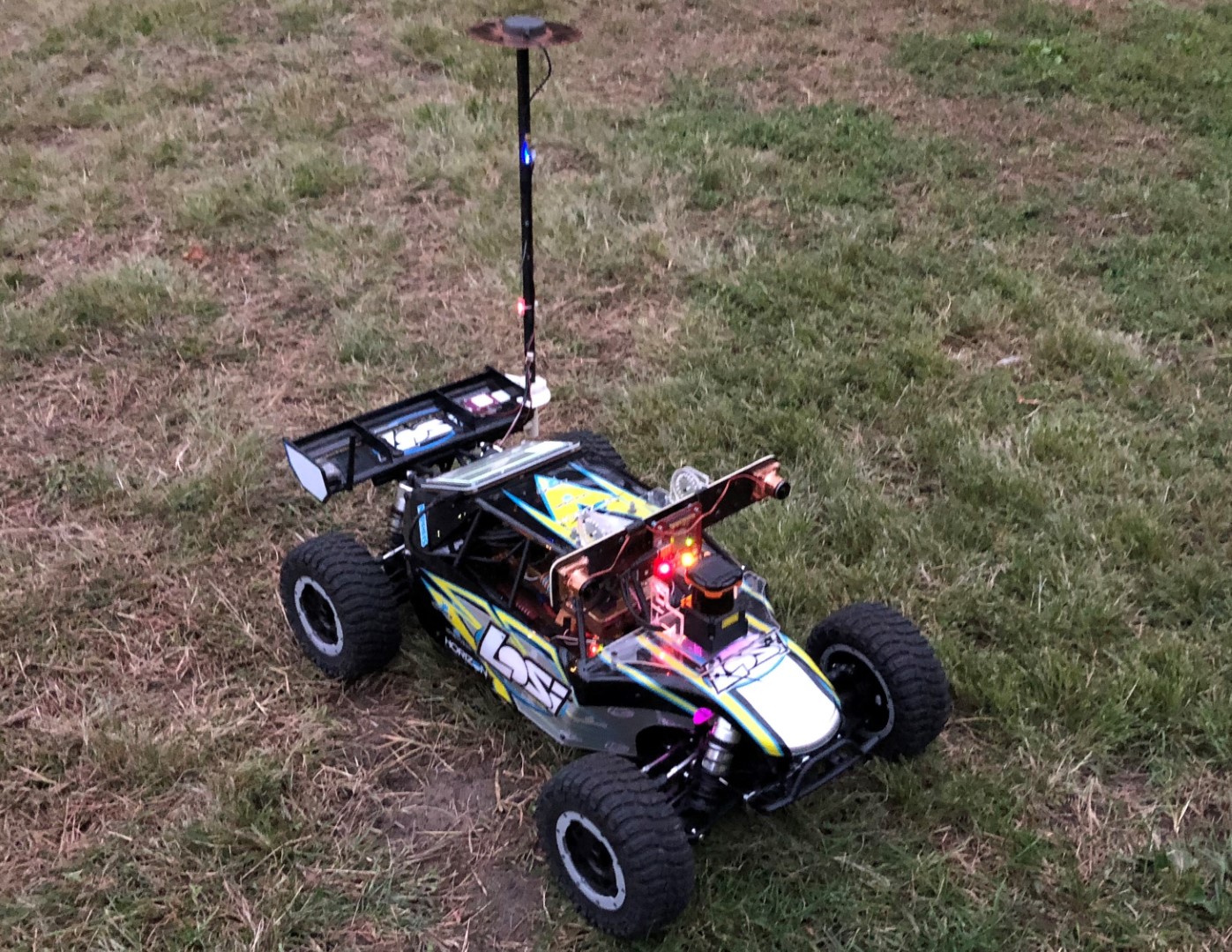}
        \caption{}
        \label{fig:ugv}
    \end{subfigure}
    \begin{subfigure}[]{0.23\textwidth}
    	\captionsetup{justification=centering}
        \includegraphics[width=\textwidth]{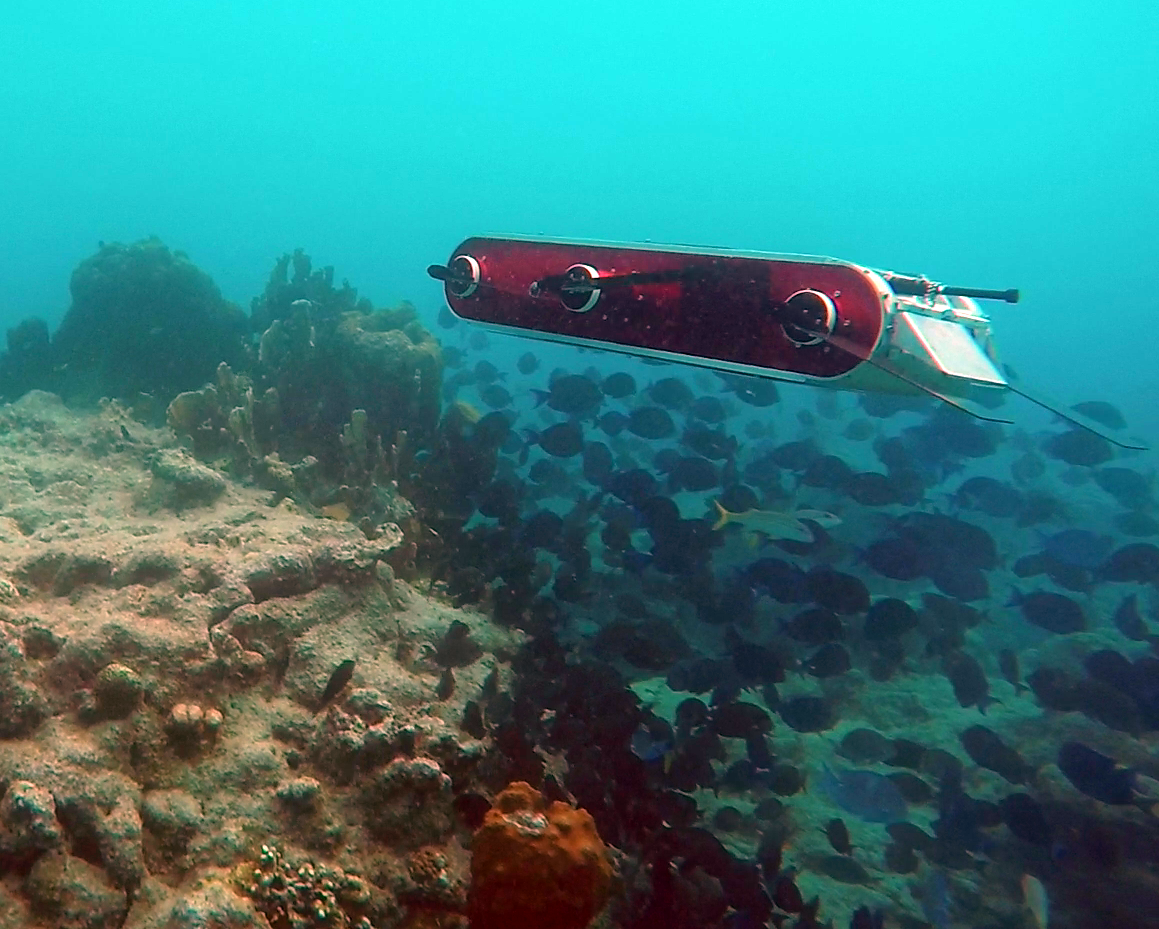}
        \caption{}
        \label{fig:aqua}
    \end{subfigure}
    \begin{subfigure}[]{0.25\textwidth}
    	\captionsetup{justification=centering}
        \includegraphics[width=\textwidth]{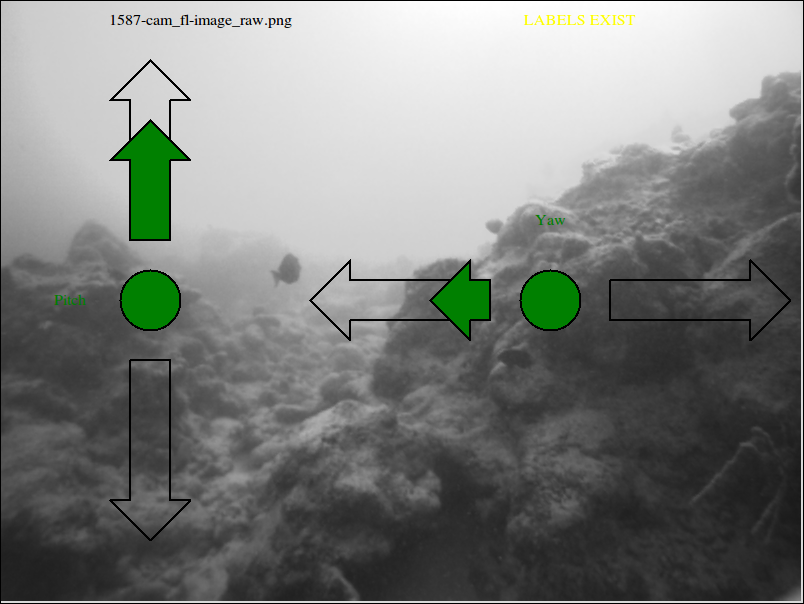}
        \caption{}
        \label{fig:low_level}
    \end{subfigure}
    \captionsetup{}
    \caption{a)~Aerial view of hot-spots where fires may spread if not extinguished, b)~initial attack crew work to extinguish hot-spots, c)~hexacopter \ac{UAV} used to take aerial photos, d)~off-road \ac{UGV} used for experiments, e)~an Aqua class robot autonomously swimming in dense coral, f)~a forward-facing image taken from the robot showing the predicted yaw and pitch steering angles.}
    \vspace{-1em}
    \label{fig:intro}
    \end{center}
\end{figure}

We propose a strategy for a team of autonomous vehicles to collaborate on information based planning for sampling dynamic environments that are dangerous for the humans. Some of the example applications include -- collecting thermal imagery in a wildfire-affected site to assist with detecting spot fires, fire mapping and monitoring fire progression (Fig.\ref{fig:initial_attack}), visual inspection and monitoring of coral reefs by planning on aerial image from an \ac{UAV} and learning to navigate (Fig.\ref{fig:low_level}) tough underwater environments with an \ac{AUV} (Fig.\ref{fig:aqua}).

Each year over 5 million hectares are burned due to forest fires in Canada and the United States. A profound number of people, ground-vehicles and aerial-vehicles are used to detect, suppress and extinguish forest fires. Wildfire suppression is a complex task that involves the dispatch of limited resources to the most at-risk fires where detection is often done through thermal imaging (Fig.\ref{fig:hotspot}) and human observations (Fig.\ref{fig:initial_attack}). Manned helicopters are used to survey areas for \textit{hot-spots} both before and after a wildfire which can spread beyond control. Several techniques are used for fire suppression, but the most notable is Initial Attack which can contain/control over 90\% of wildfires\footnote{\url{https://www.nifc.gov/PUBLICATIONS/redbook/2000/Chapter9.pdf}}. Several factors determine the initial attack strategy including:
\begin{enumerate}
	\item location
	\item proximity to human-life and property
	\item fuel type (i.e. trees or grass)
	\item terrain topography (flatness, rivers, mountain slope or valley)
	\item weather conditions
	\item fire size (height and area)
\end{enumerate}

\begin{figure*}[h]
    \begin{center}
    \captionsetup{justification=centering}
    \includegraphics[width=0.9\textwidth]{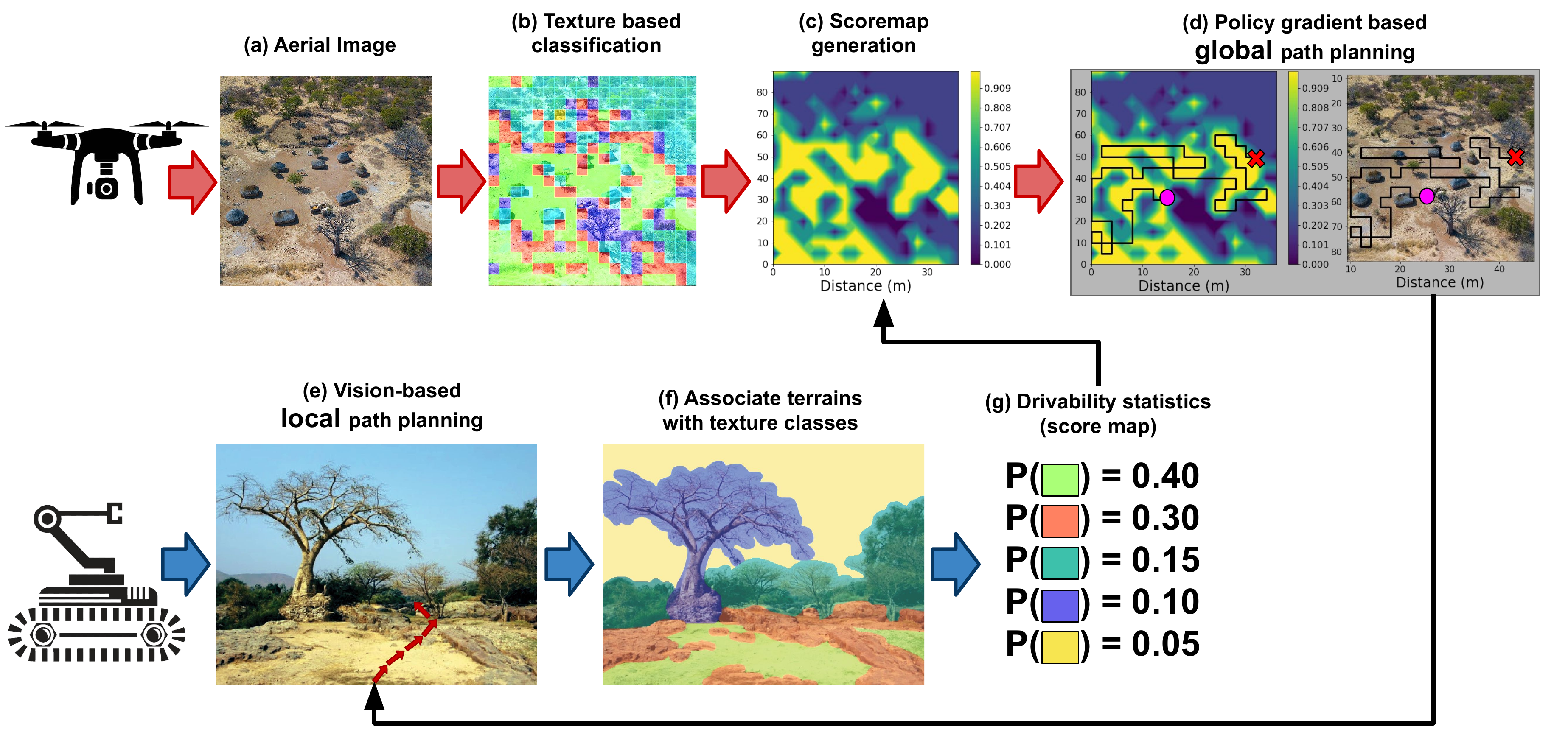}
    \captionsetup{justification=centering}
    \caption{Overview of our approach}
    \label{fig:flowchart}
    \vspace{-2em}
    \end{center}
\end{figure*}

Initial Attack techniques include extinguishing fires directly by firefighters using hand tools for small fires such as those under one meter or smouldering in the ground. Slightly larger fires which are too intense for firefighters are directly extinguished using bulldozers and fire engines or by dropping retardant. For large fires that cannot be attacked directly, a perimeter of natural (such as roads) or artificial barriers (created by limiting fuel sources) are used to control the fire and is allowed to burn out.


\vspace{1em}
Recent advances in aerial and ground robotics could one day be used to assist in the detection and suppression of fires \cite{beachly2018fire, torresan2017forestry}. Low-cost \acp{UAV} can replace much of the current work being done by manned helicopters by carrying thermal cameras to detect \textit{hot-spots} at a high altitude as well as take local temperature measurements on the ground where accessible. Further, \acp{UGV} can be used to perform in situ measurements, digging up hot-spots, extinguishing by spraying retardant, bulldozing fuel sources to create perimeters and even supplying resources to manned crews. In literature, collaborative teams of \acp{UAV} and \acp{UGV}  have been used for localizing each other \cite{vaughan2000fly}, or for shared coverage, or in search and rescue scenarios or for target tracking \cite{yu2011probabilistic} and surveillance \cite{grocholsky2006cooperative}. We focus this work on using an heterogeneous robotic team for coordinated information gathering in outdoor environments.

In this work, we propose a methodology for global and local path planning of an \ac{UGV} using imagery from an aerial vehicle. We seek to generate efficient paths for the \ac{UGV} by segmenting the aerial image based on textures and labeling each class with a score associated to its \textit{drivability}. As the ground-vehicle navigates, the performance in terms of progress towards the destination and incidence of obstacles measures the drivability of the terrain class and the drivability score is updated. 
Robots can play an important role in the process through aerial thermal imaging. In this work we focus on the global and local planning for an \ac{UGV} for sampling applications. The presented methodology of collaborative sampling can find use in various real world applications, such as search and rescue operations, sampling large-scale environmental scientific data, and large-scale coral reef monitoring using an \ac{UAV} and an \ac{AUV}.

\section{Our Approach}

The flowchart in Fig.\ref{fig:flowchart} presents an overview of our strategy for collaborative coverage and sampling on an outdoor scene. We split our approach into two planning phases: \emph{Global planning}, where the high-level planner chooses locations that need to be visited to achieve good sampling and plans an efficient path (encoded as a discrete set of waypoints) to visit these locations; and \emph{Local planner}, which handles short-term navigation between the waypoints (preferring good drivable terrains) and small deviations caused by obstacles.

A collaborative phase is responsible for labeling the terrain classes (used by the global planner) with the drivability feedback from the \ac{UGV}. This update will result in an improved global plan and re-planning is necessary when the local planner on the \ac{UGV} is unable to circumnavigate an obstacle. We discuss the building blocks of our system in the following subsections and later present some preliminary results from the evaluation of these building blocks.

\subsection{Texture classification and learning the scoremap}

\begin{figure}[h]
    \begin{center}
    \begin{subfigure}[]{0.48\textwidth}
    	\captionsetup{justification=centering}
        \includegraphics[width=\textwidth]{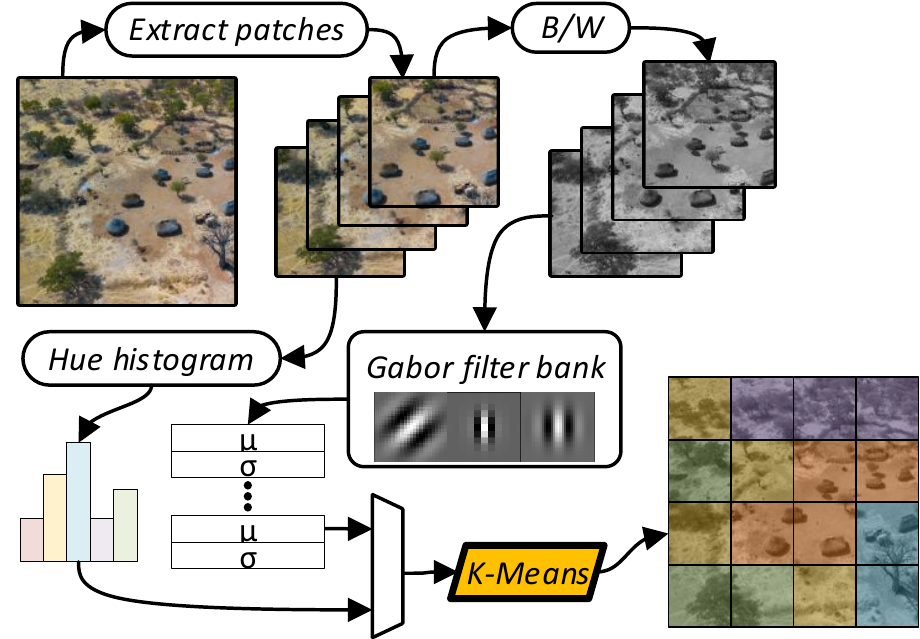}
        \caption{}
        \label{fig:seg-flow}
    \end{subfigure}
    \begin{subfigure}[]{0.24\textwidth}
    	\captionsetup{justification=centering}
        \includegraphics[width=\textwidth]{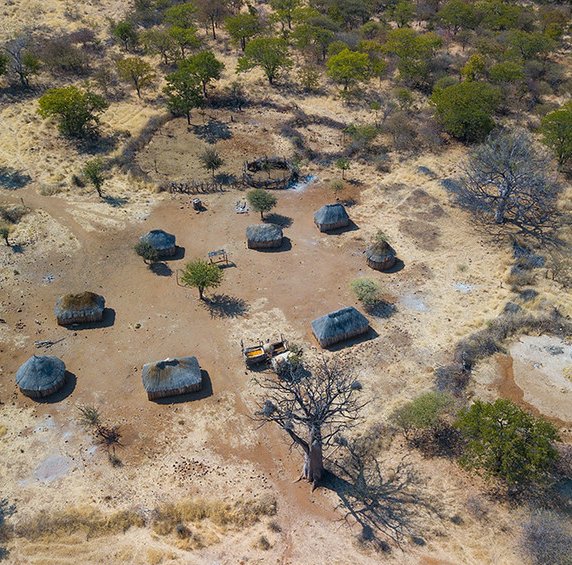}
        \caption{}
        \label{fig:aerial}
    \end{subfigure}
    \begin{subfigure}[]{0.24\textwidth}
    	\captionsetup{justification=centering}
        \includegraphics[width=\textwidth]{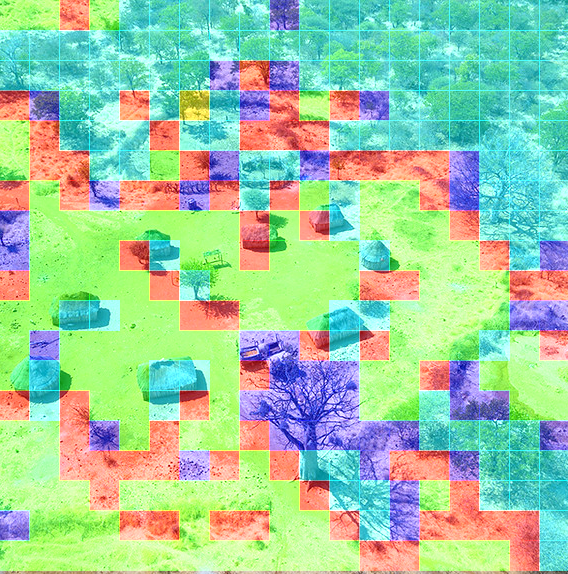}
        \caption{}
        \label{fig:sample_seg}
    \end{subfigure}
    \captionsetup{}
    \caption{a)~Segmentation pipeline, (b)~Aerial image from an \ac{UAV}, (c)~Segmented aerial image based on a textural classifier.}
    \label{fig:preliminary_results2}
    \vspace{-1em}
    \end{center}
\end{figure}

Our texture classification (as shown in Fig.\ref{fig:seg-flow}) classifies patches of the image into a number of distinct texture classes. The texture classifier uses several Gabor filters to describe the texture \cite{fogelSagiGaborTexture1989}. The filters are convolved with the image to produce robust energy statistics, and several filters allow the classifier to be rotation and scale invariant. 

The distribution of hue content is parameterized as a histogram. The hue histogram parameters and Gabor order statistics are combined into a feature vector, which are fed into a K-means classifier to discriminate between a preset number of texture classes. Fig.\ref{fig:sample_seg} illustrates the texture based segmentation on an aerial image in Fig.\ref{fig:aerial} using our classification pipeline.

Independent of the texture classifier, we label each texture class with a drivability score which is learned and updated on-line using feedback from the performance of the \ac{UGV} (Fig. \ref{fig:flowchart}e, f, and g), such as rate of progress towards goal, incidence of obstacles, and terrain topology and rugosity.


\subsection{Policy Gradient based global path planning}

We use policy search on aggregated state space \cite{manjanna2018iser} to generate paths for the \ac{UGV} such that the drivable regions are covered. In this approach \cite{manjanna2018reinforcement} a continuous two-dimensional sampling region is discretized into uniform grid-cells, such that the robot's position $\robotloc$ can be represented by a pair of integers $\robotloc \in \mathbb{Z}^2$. Each grid-cell $(i,j)$ is assigned a score $\rmap(i,j)$ indicating the expected drivabillity in that cell. The goal is to maximize the total accumulated score $J$ over a trajectory $\tau$ within a fixed amount of time $T$. To specify the robot's behavior we use a parametrized policy $\pi_{\vec{\theta}}(\vec{s},\vec{a}) = p(\vec{a}|\vec{s};\vec{\theta})$ that maps the current state $\vec{s}$ of sampling to a distribution over possible \emph{actions} $\vec{a}$. The aim here is to automatically find good parameters $\vec{\theta}$, after which the policy can be deployed without additional training on new problems. We use a multi-resolution feature representation centered around the robot as explained in \cite{manjanna2018iser}. In this representation, the feature cells grow in size along with the distance from the robot. This results in high resolution features close to the robot and lower resolution features further from the robot's current position. 

We use the GPOMDP and REINFORCE algorithms \cite{baxter2001infinite,sutton2000policy,deisenroth2013survey,kober2013reinforcement} for computing the policy gradient as they have fewer hyper-parameters, making it easier to deploy, 
\begin{multline*}
\label{eq:gradient}
\nabla_\theta J_\theta = \frac{1}{m}\sum_{i=1}^m\sum_{t=0}^{H-1}\nabla_\theta \log \pi_\theta(a_t^{(i)}|s_t^{(i)})\\
\left(\sum_{j=t}^{H-1}r(s_j^{(i)},a_j^{(i)}) - b(s_t^{(i)})\right).
\end{multline*}
In this equation, the gradient is based on $m$ sampled trajectories with horizon $H$ and state $\vec{s}_j^{(i)}$ at the $j^{\text{th}}$ time-step of the $i^{\text{th}}$ sampled roll-outs. Furthermore, $b$ is a variance-reducing baseline. In our experiments, we set the baseline to the observed average reward. 

The path planned on the generated scoremap is presented in Fig. \ref{fig:flowchart}(d). Fig. \ref{fig:flowchart}(d) also illustrates the planned path overlayed on the initial aerial image.

\subsection{Vision based local path planning for \ac{UGV}}
The \ac{UGV} uses vision-based navigation which steers the vehicle towards the next way-point while avoiding obstacles and preferring paths which are good for driving and provide good texture for visual state estimation. 

A \ac{CNN} is used to predict steering angles that avoid obstacles and leads to driving on good terrain. To train the \ac{CNN}, we perform behavioural cloning not unlike the DAGGER algorithm\cite{ross2011reduction}. The \ac{UGV} is remotely controlled in both good and poor configurations such as very close to and far from obstacles. A Resnet-18 based \ac{CNN} is trained to predict steering angle $\theta$ from a set C=$\lbrace-M,...,0,...,M,\rbrace$, where $M$ is a discrete number of steering angles in each the left and right direction. From the data collected in the remotely controlled step, a set of images and one-hot encoded steering angles make up a dataset $D$. The cost function for training the network is as follows:
\begin{align*}
\begin{split}
\mathcal{L}\left(D,\mathbf{w}\right) &= \mathcal{L}_{\mathrm{pred}}\left(D,\mathbf{w}\right) + \lambda_1\mathcal{L}_{\mathrm{reg}}\left(D,\mathbf{w}\right)\\
\mathcal{L}_{\mathrm{pred}} &= \sum_{(\mathbf{x}_i, \mathbf{\theta}_i) \in D} l(\mathbf{f}^{(\theta)}_i, \mathbf{\theta}_i, \mathbf{w})
\end{split}
\end{align*}
 where the regularization term corresponds to the KL-divergence term in~\cite{Gal2017ConcreteB}. The prediction loss for yaw steering actions was given by
\begin{align*}
\begin{split}
l(\mathbf{f}^{(\theta)}_i, \mathbf{\theta}_i, \mathbf{w}) &= -\sum_j \theta_{ij} \log f^{(\theta)}_{ij} - \lambda_2 \sum_j f^{(\theta)}_{ij} \log f^{(\theta)}_{ij}
\end{split}
\end{align*}
where first term corresponds to the cross-entropy between the network predictions and the smoothed labels, and the second term is the penalty for overconfident predictions, which aims to maximize the entropy of the predictive distributions. The hyper-parameters $\lambda_1$ and $\lambda_2$ determine the weights of the KL-divergence regularization and the entropy penalty and are selected manually.



\begin{figure*}[h]
    \begin{center}
    \begin{subfigure}[]{0.2\textwidth}
    	\captionsetup{justification=centering}
        \includegraphics[width=\textwidth]{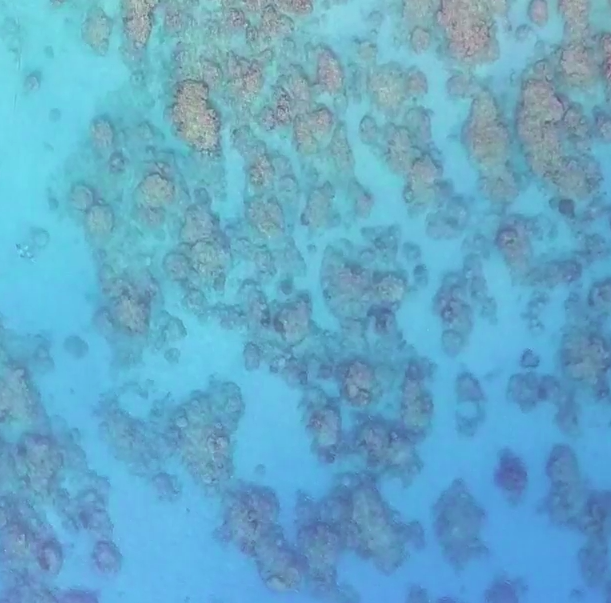}
        \caption{}
        \label{fig:aerial_reef}
    \end{subfigure}
    \begin{subfigure}[]{0.25\textwidth}
    	\captionsetup{justification=centering}
        \includegraphics[width=\textwidth]{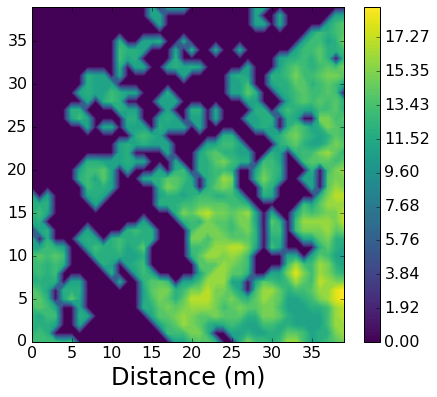}
        \caption{}
        \label{fig:scoremap_test2}
    \end{subfigure}
    \begin{subfigure}[]{0.22\textwidth}
    	\captionsetup{justification=centering}
        \includegraphics[width=\textwidth]{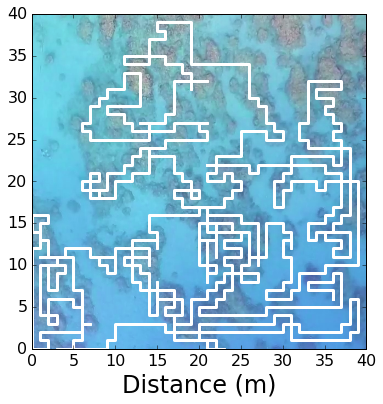}
        \caption{}
        \label{fig:reef_pg}
    \end{subfigure}
    \begin{subfigure}[]{0.24\textwidth}
    	\captionsetup{justification=centering}
        \includegraphics[width=\textwidth]{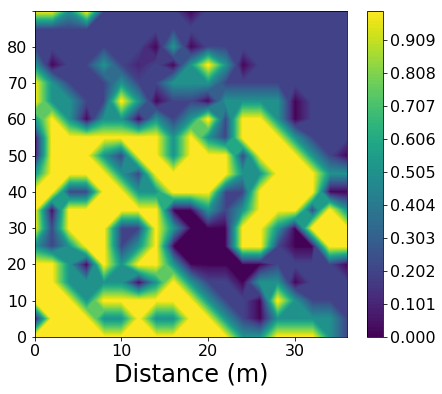}
        \caption{}
        \label{fig:scoremap}
    \end{subfigure}
    \begin{subfigure}[]{0.25\textwidth}
    	\captionsetup{justification=centering}
        \includegraphics[width=\textwidth]{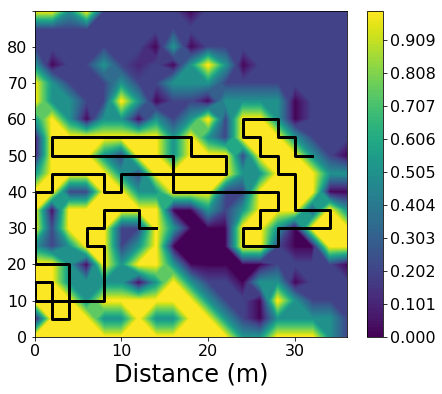}
        \caption{}
        \label{fig:path_scoremap}
    \end{subfigure}
    \begin{subfigure}[]{0.23\textwidth}
    	\captionsetup{justification=centering}
        \includegraphics[width=\textwidth]{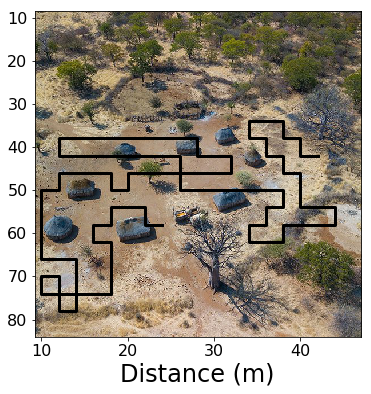}
        \caption{}
        \label{fig:path_sample}
    \end{subfigure}
    \begin{subfigure}[]{0.5\textwidth}
    	\captionsetup{justification=centering}
        \includegraphics[width=\textwidth]{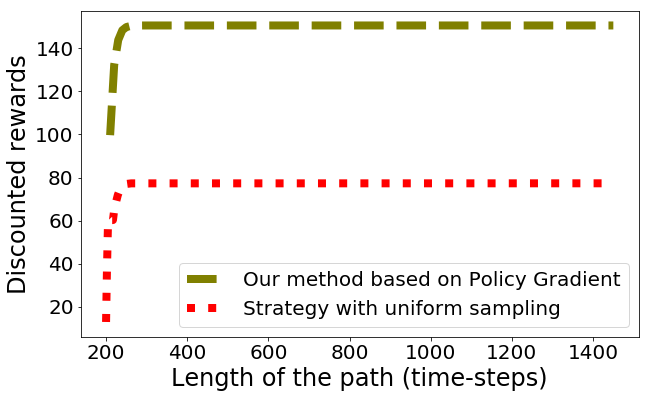}
        \caption{}
        \label{fig:pg_result}
    \end{subfigure}
    \captionsetup{}
    \vspace{1em}
    \caption{a)~Aerial view of a coral reef,  b)~Scoremap generated using texture information in the aerial image,  c)~Trajectory using the policy gradient search based planner on the scoremap, d)~Scoremap generated using the segmentations from Fig.\ref{fig:sample_seg}, e)~The path planned using the policy gradient based planner, f)~The path generated is overlaid on top of the initial aerial image, g)~Plot comparing the total discounted rewards collected by our strategy and the uniform sampling strategy.}
    \label{fig:preliminary_results}
    \end{center}
\end{figure*}

\section{Preliminary Results}

We evaluated the building blocks of our system on both marine and terrestrial setups. In the marine example, the application is to cooperatively collect better visual samples of the coral reefs in a limited time budget using the \ac{UAV} for global planning and an \ac{AUV} for local planning and execution. The presented results are from experiments conducted in field at Folkestone Marine Reserve in Barbados, over a region known to have several coral outcrops. The \ac{AUV} used in these experiments is a Aqua class underwater robot \cite{sattar2008enabling}. The \ac{AUV} has a vision based local planner to swim between reefs collecting high-resolution visual samples. We used the aerial images from the \ac{UAV} to plan trajectories for the \ac{AUV} (as presented in Fig.\ref{fig:aerial_reef}, Fig.\ref{fig:scoremap_test2}, and Fig.\ref{fig:reef_pg}), such that the information gain about the coral reefs is maximized with a limited time budget. In the terrestrial example, we used the aerial images from the \ac{UAV} presented in Fig.\ref{fig:aerial} to generate a scoremap indicating the drivability of the terrain as illustrated in Fig.\ref{fig:scoremap}. The policy gradient based path planner then plans a trajectory for the \ac{UGV} (as presented in Fig.\ref{fig:path_scoremap} and Fig.\ref{fig:path_sample}), such that the coverage of drivable terrain is maximized with a limited time.

In one of our recent work \cite{manjanna2018iser}, we evaluated our policy gradient based path planner in the marine domain to monitor coral reefs. The results presented in Fig.\ref{fig:pg_result} illustrate that we are able to achieve higher discounted rewards in comparison to a uniform sampling technique. The goal here is to maximize the discounted rewards as the time budget is limited for the robotic vehicle.

\begin{figure*}[h]
    \begin{center}
    	\captionsetup{justification=centering}
        \includegraphics[width=\linewidth]{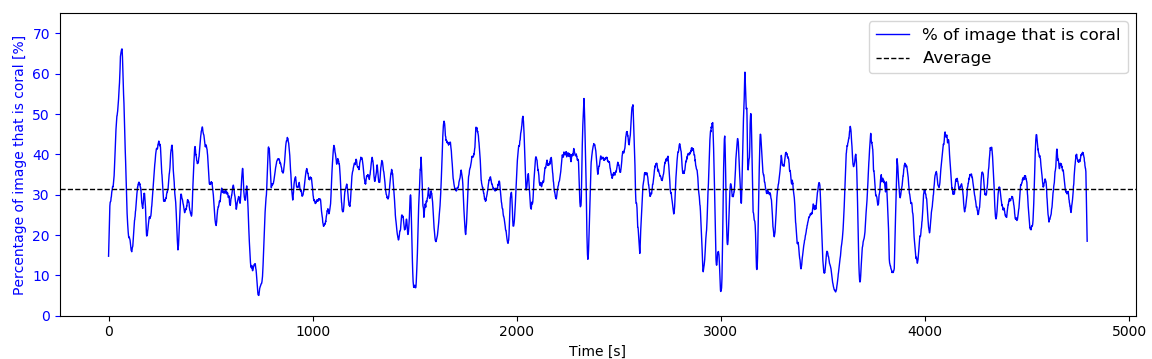}
        \caption{The percentage of image containing coral. It's expected that at most 50\% of the image can be coral since the top half of the image is \textit{normally} above the horizon.}
        \label{fig:coral-image-coverage}
    \end{center}
\end{figure*}

We have performed preliminary validation of our local planner in the same marine environment as we validated the global planner. In \cite{Manderson2018iser}, \cite{Manderson2018iros} we performed long-distance autonomous underwater navigation in close-proximity to coral reefs (on average 43 cm from coral) without collision. We manually labelled approximately 13,000 images taken in varying configurations and several lighting configurations with expected yaw and pitch and used them to train a \ac{CNN}. Our experimental validation resulted in over one km of collision-free navigation in the open sea, while maintaining good coverage of coral which was our target observation. Fig. \ref{fig:coral-image-coverage} shows the percentage of the forward-facing images that were covered with coral, showing that our robot maintained consistent coverage of the target terrain (coral). On average, 33\% of the image showed the presence of coral indicating the effectiveness of navigating coral region as opposed to barren regions. Since the horizon is located in the center of image, we expect the coral in view to be much lower than 50\%. In an upcoming field trial, we plan to evaluate all the building blocks and the whole pipeline. 

\begin{figure}[tbh]
    \begin{center}

    	\captionsetup{justification=centering}
        \includegraphics[width=\linewidth]{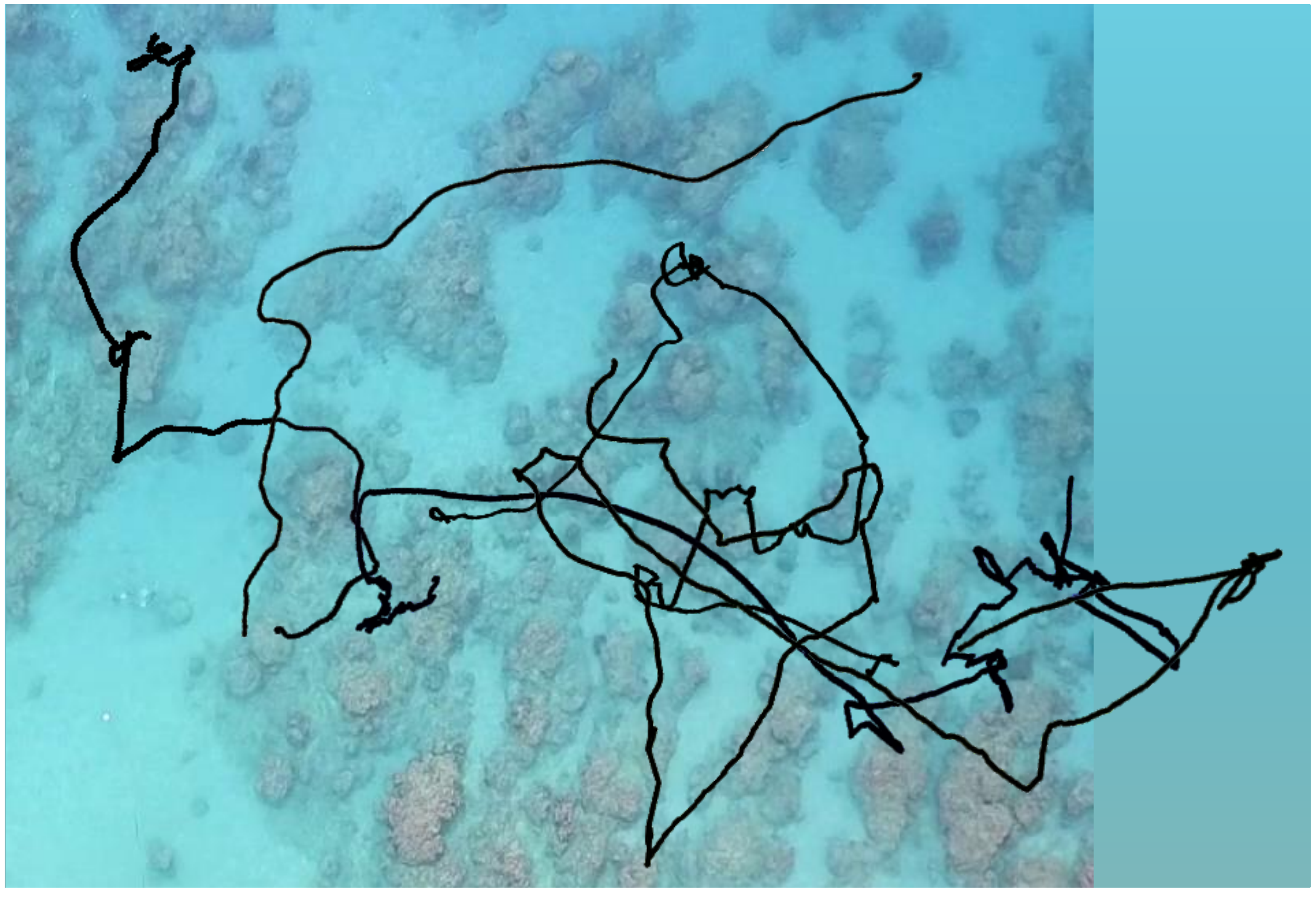}
        \caption{The final paths swam by the Aqua robot covering over 1 km.}
        \label{fig:reef}

    \end{center}
\end{figure}

\section{Conclusion} 
\label{sec:conclusion}
In this paper we have outlined a system for cooperative path planning using information acquired from an \ac{UAV} to plan paths for an \ac{UGV}. Drivability performance on the \ac{UGV} is used to update a scoremap to plan the path iteratively. We demonstrate the effectiveness of the subsystems in our preliminary experiments in a marine domain with collaboration between an aerial and an underwater vehicles. In our current and future work, we are tightening the interaction between the \ac{UAV} and \ac{UGV} along with a unified sampling goal. We expect that this work will lead to a robust and generalized framework for deploying cooperative aerial and ground robots in complex, unstructured and dynamic environments. The active-learning framework allows the system to work from previously seen environments, but also incorporate new unseen information into the model for improved planning and execution.

\begin{acronym}
    \acro{UAV}{Unmanned Aerial Vehicle}
	\acro{UGV}{Unmanned Ground Vehicle}	
	\acro{CNN}{Convolutional Neural Network}
	\acro{AUV}{Autonomous Underwater Vehicle}
\end{acronym}

\bibliographystyle{plainnat}
\bibliography{heterogeneous_robot_teams}

\begin{thebibliography}{17}
\providecommand{\natexlab}[1]{#1}
\providecommand{\url}[1]{\texttt{#1}}
\expandafter\ifx\csname urlstyle\endcsname\relax
  \providecommand{\doi}[1]{doi: #1}\else
  \providecommand{\doi}{doi: \begingroup \urlstyle{rm}\Url}\fi

\bibitem[Baxter and Bartlett(2001)]{baxter2001infinite}
Jonathan Baxter and Peter~L Bartlett.
\newblock Infinite-horizon policy-gradient estimation.
\newblock \emph{Journal of Artificial Intelligence Research}, 15:\penalty0
  319--350, 2001.

\bibitem[Beachly et~al.(2018)Beachly, Detweiler, Elbaum, Duncan, Hildebrandt,
  Twidwell, and Allen]{beachly2018fire}
Evan Beachly, Carrick Detweiler, Sebastian Elbaum, Brittany Duncan, Carl
  Hildebrandt, Dirac Twidwell, and Craig Allen.
\newblock Fire-aware planning of aerial trajectories and ignitions.
\newblock In \emph{2018 IEEE/RSJ International Conference on Intelligent Robots
  and Systems (IROS)}, pages 685--692. IEEE, 2018.

\bibitem[Deisenroth et~al.(2013)Deisenroth, Neumann, and
  Peters]{deisenroth2013survey}
Marc~Peter Deisenroth, Gerhard Neumann, and Jan Peters.
\newblock A survey on policy search for robotics.
\newblock \emph{Foundations and Trends{\textregistered} in Robotics},
  2\penalty0 (1--2):\penalty0 1--142, 2013.

\bibitem[Fogel and Sagi(1989)]{fogelSagiGaborTexture1989}
I.~Fogel and D.~Sagi.
\newblock Gabor filters as texture discriminator.
\newblock \emph{Biological Cybernetics}, 61\penalty0 (2):\penalty0 103--113,
  1989.
\newblock ISSN 0340-1200.
\newblock \doi{10.1007/BF00204594}.
\newblock URL \url{http://dx.doi.org/10.1007/BF00204594}.

\bibitem[Gal et~al.(2017)Gal, Hron, and Kendall]{Gal2017ConcreteB}
Yarin Gal, Jiri Hron, and Alex Kendall.
\newblock {Concrete Dropout}.
\newblock In \emph{Advances in Neural Information Processing Systems 30
  (NIPS)}, 2017.

\bibitem[Grocholsky et~al.(2006)Grocholsky, Keller, Kumar, and
  Pappas]{grocholsky2006cooperative}
Ben Grocholsky, James Keller, Vijay Kumar, and George Pappas.
\newblock Cooperative air and ground surveillance.
\newblock \emph{IEEE Robotics \& Automation Magazine}, 13\penalty0
  (3):\penalty0 16--25, 2006.

\bibitem[Kober et~al.(2013)Kober, Bagnell, and Peters]{kober2013reinforcement}
Jens Kober, J~Andrew Bagnell, and Jan Peters.
\newblock Reinforcement learning in robotics: A survey.
\newblock \emph{The International Journal of Robotics Research}, 32\penalty0
  (11):\penalty0 1238--1274, 2013.

\bibitem[Manderson et~al.(2018{\natexlab{a}})Manderson, Cheng, Meger, and
  Dudek]{Manderson2018iser}
Travis Manderson, Ran Cheng, Dave Meger, and Gregory Dudek.
\newblock {Navigation in the Service of Enhanced Pose Estimation}.
\newblock In \emph{International Symposium on Experimental Robotics (ISER)},
  2018{\natexlab{a}}.

\bibitem[Manderson et~al.(2018{\natexlab{b}})Manderson, Higuera, Cheng, and
  Dudek]{Manderson2018iros}
Travis Manderson, Juan Camilo~Gamboa Higuera, Ran Cheng, and Gregory Dudek.
\newblock {Vision-based autonomous underwater swimming in dense coral for
  combined collision avoidance and target selection}.
\newblock In \emph{2018 IEEE/RSJ International Conference on Intelligent Robots
  and Systems (IROS)}, pages 1885--1891. IEEE, 2018{\natexlab{b}}.

\bibitem[Manjanna et~al.(2018{\natexlab{a}})Manjanna, van Hoof, and
  Dudek]{manjanna2018iser}
Sandeep Manjanna, Herke van Hoof, and Gregory Dudek.
\newblock Policy search on aggregated state space for active sampling.
\newblock In \emph{2018 International Symposium on Experimental Robotics
  (ISER)}, pages 1--7, 2018{\natexlab{a}}.

\bibitem[Manjanna et~al.(2018{\natexlab{b}})Manjanna, van Hoof, and
  Dudek]{manjanna2018reinforcement}
Sandeep Manjanna, Herke van Hoof, and Gregory Dudek.
\newblock Reinforcement learning with non-uniform state representations for
  adaptive search.
\newblock In \emph{2018 IEEE International Symposium on Safety, Security, and
  Rescue Robotics (SSRR)}, pages 1--7. IEEE, 2018{\natexlab{b}}.

\bibitem[Ross et~al.(2011)Ross, Gordon, and Bagnell]{ross2011reduction}
St{\'e}phane Ross, Geoffrey Gordon, and Drew Bagnell.
\newblock A reduction of imitation learning and structured prediction to
  no-regret online learning.
\newblock In \emph{Proceedings of the fourteenth international conference on
  artificial intelligence and statistics}, pages 627--635, 2011.

\bibitem[Sattar et~al.(2008)Sattar, Dudek, Chiu, Rekleitis, Giguere, Mills,
  Plamondon, Prahacs, Girdhar, Nahon, et~al.]{sattar2008enabling}
Junaed Sattar, Gregory Dudek, Olivia Chiu, Ioannis Rekleitis, Philippe Giguere,
  Alec Mills, Nicolas Plamondon, Chris Prahacs, Yogesh Girdhar, Meyer Nahon,
  et~al.
\newblock Enabling autonomous capabilities in underwater robotics.
\newblock In \emph{{IEEE/RSJ International Conference on Intelligent Robots and
  Systems}}, pages 3628--3634, 2008.

\bibitem[Sutton et~al.(2000)Sutton, McAllester, Singh, and
  Mansour]{sutton2000policy}
Richard~S Sutton, David~A McAllester, Satinder~P Singh, and Yishay Mansour.
\newblock Policy gradient methods for reinforcement learning with function
  approximation.
\newblock In \emph{Advances in neural information processing systems}, pages
  1057--1063, 2000.

\bibitem[Torresan et~al.(2017)Torresan, Berton, Carotenuto, Di~Gennaro, Gioli,
  Matese, Miglietta, Vagnoli, Zaldei, and Wallace]{torresan2017forestry}
Chiara Torresan, Andrea Berton, Federico Carotenuto, Salvatore~Filippo
  Di~Gennaro, Beniamino Gioli, Alessandro Matese, Franco Miglietta, Carolina
  Vagnoli, Alessandro Zaldei, and Luke Wallace.
\newblock Forestry applications of uavs in europe: A review.
\newblock \emph{International Journal of Remote Sensing}, 38\penalty0
  (8-10):\penalty0 2427--2447, 2017.

\bibitem[Vaughan et~al.(2000)Vaughan, Sukhatme, Mesa-Martinez, and
  Montgomery]{vaughan2000fly}
Richard~T Vaughan, Gaurav~S Sukhatme, Francisco~J Mesa-Martinez, and James~F
  Montgomery.
\newblock Fly spy: Lightweight localization and target tracking for cooperating
  air and ground robots.
\newblock In \emph{Distributed autonomous robotic systems 4}, pages 315--324.
  Springer, 2000.

\bibitem[Yu et~al.(2011)Yu, Beard, Argyle, and
  Chamberlain]{yu2011probabilistic}
Huili Yu, Randal~W Beard, Matthew Argyle, and Caleb Chamberlain.
\newblock Probabilistic path planning for cooperative target tracking using
  aerial and ground vehicles.
\newblock In \emph{Proceedings of the 2011 American Control Conference}, pages
  4673--4678. IEEE, 2011.

\end{thebibliography}

\end{document}